\documentclass[letterpaper, 10 pt, conference]{ieeeconf}  

\IEEEoverridecommandlockouts                              

\usepackage{graphics} 
\usepackage{epsfig} 

\usepackage{mathtools}
\usepackage{amsmath} 
\usepackage{amssymb}  
\usepackage{algorithm}
\usepackage{algorithmic}
\usepackage{todonotes}
\usepackage{xcolor}
\usepackage{graphicx}
\usepackage{caption,subcaption}

\usepackage{multirow}
\usepackage{mathrsfs}
\usepackage{ctable} 
\usepackage{url}
\usepackage{array,booktabs,ragged2e}
\usepackage[english]{babel}
\usepackage{gensymb}
\usepackage{hhline}
\usepackage{hyperref}
\usepackage{cite}

\newtheorem{problem}{Problem}
\newtheorem{lemma}{Lemma}
\newtheorem{proposition}{Proposition}

\renewcommand{\r}{{\mathbf r}}

\newcommand{\z}{{\mathbf z}}

\newcommand{\R}{\mathbb{R}}

\newcommand{\Z}{{\mathcal Z}}

\newcommand{\x}{\mathbf{x}}

\renewcommand{\v}{{\mathbf v}}

\newcommand{\p}{{\mathbf p}}

\newcommand{\VV}{{\mathbf V}}
\newcommand{\YY}{{\mathbf Y}}
\newcommand{\ZZ}{{\mathbf Z}}

\newcommand{\expc}[1]{ {\{ #1 \} } }
\newcommand{\expct}[1]{ {\{ #1 \}^T } }
\newcommand{\expe}[1]{ {\otimes #1 } }
\newcommand{\expet}[1]{ {{\otimes #1}^T} }

\renewcommand{\L}{\mathcal{L}}

\usepackage{tikz}

\newcommand\copyrighttext{%
  \footnotesize \textcopyright 2023 IEEE.  Personal use of this material is permitted.  Permission from IEEE must be obtained for all other uses, in any current or future media, including reprinting/republishing this material for advertising or promotional purposes, creating new collective works, for resale or redistribution to servers or lists, or reuse of any copyrighted component of this work in other works.}
\newcommand\copyrightnotice{%
\begin{tikzpicture}[remember picture,overlay]
\node[anchor=south,yshift=10pt] at (current page.south) {\fbox{\parbox{\dimexpr\textwidth-\fboxsep-\fboxrule\relax}{\copyrighttext}}};
\end{tikzpicture}%
}

\title{\LARGE \bf
Visibility-Constrained Control of Multirotor via Reference Governor }

\author{Dabin Kim, Matthias Pezzutto, Luca Schenato, and H. Jin Kim
\thanks{This research was supported by Unmanned Vehicles Core Technology Research and Development Program through the National Research  Foundation of Korea(NRF) and Unmanned Vehicle Advanced Research Center(UVARC) funded by the Ministry of Science and ICT, the Republic of Korea(NRF-2020M3C1C1A010864)}
\thanks{Dabin Kim and H. Jin Kim are with the Department of Aerospace Engineering, Seoul National University, Seoul 08826, South Korea (e-mail: dabin404@snu.ac.kr, hjinkim@snu.ac.kr).}
\thanks{Matthias Pezzutto is with the Service d'Automatique et d'Analyse des Systèmes (SAAS) of the Université Libre De Bruxelles, Bruxelles, Belgium. (email: matthias.pezzutto@ulb.be)}
\thanks{Luca Schenato is with the Department of Information Engineering of the  University of Padova, Padova, Italy (email: schenato@dei.unipd.it)
}
}

\begin{document}

\maketitle
\copyrightnotice
\thispagestyle{empty}
\pagestyle{empty}

\vspace{-3.5mm}
\begin{abstract}
For safe vision-based control applications, perception-related constraints have to be satisfied in addition to other state constraints. In this paper, we deal with the problem where a multirotor equipped with a camera needs to maintain the visibility of a point of interest while tracking a reference given by a high-level planner. 
We devise a method based on reference governor that, differently from existing solutions, is able to enforce control-level visibility constraints with theoretically assured feasibility. To this end, we design a new type of reference governor for linear systems with polynomial constraints which is capable of handling time-varying references.
The proposed solution is implemented online for the real-time multirotor control with visibility constraints and validated with simulations and an actual hardware experiment.
\end{abstract}

\section{Introduction}

Multirotors equipped with a camera have been popular in a large number of applications, such as navigation \cite{kim2021topology}, inspection \cite{paneque2022perception}, and visual servoing \cite{zheng2019toward}.
In this kind of applications, the multirotor has to achieve a given high-level objective while keeping a point or region of interest inside the Field of View (FoV) of the camera.
In fact, a loss of visibility during the mission may be irrecoverable or can even compromise safety.

The loss of visibility can occur when other control objectives are prioritized over the visibility requirements, for instance, when the desired trajectory leads to aggressive maneuvers that drive the point-of-interest (PoI) out of the FoV. Even explicit consideration of visibility in the control design may not prevent its loss if multirotor dynamics or other state limitations are not appropriately considered. To address these issues, visibility requirements can be formalized as a constraint and a constrained control method can be used to track the desired trajectory while enforcing visibility.

In the field of visual servoing, \cite{zhang2021robust} suggests a method based on  nonlinear Model Predictive Control (MPC) while \cite{zheng2019toward} develops a control barrier function (CBF) to ensure visibility with visual servo control.
These solutions are limited to visual servoing, whose goal is to regulate or drive the image feature to the desired point, 
and cannot be easily generalized for arbitrary trajectory tracking control.

For more general tracking problems with multirotors,
the minimum-time trajectory generation problem subject to visibility constraints is tackled in \cite{penin2017vision}, while \cite{falanga2018pampc} incorporates the visibility requirements as a soft constraint by designing an additional objective function in MPC. In \cite{lee2020aggressive}, feature pixels are included into the system's state and visibility is represented as state constraints. 
In these methods, the visibility constraint is addressed during the trajectory generation, while a separate tracking controller handles the low-level control. However, this approach may result in visibility constraint violations if the tracking controller fails to ensure feasibility.
A nonlinear MPC method is proposed in \cite{jacquet2020motor} to track a reference while enforcing visibility constraints, but no formal guarantees on the recursive feasibility are given. 

In this paper, we present a rigorous solution based on Reference Governor (RG) for visibility-constrained control of multirotors.
RG is an add-on control scheme which enforces constraints on a pre-stabilized system by modifying whenever necessary the reference input \cite{garone2017reference}.
More specifically, for every time instant $t$, the input value $v(t)$ is determined by solving an optimization problem. This optimization problem minimizes the distance between the new input $v(t)$ and the desired reference $r(t),$ while ensuring that the predicted system evolution, 
initiated at the current state $x(t)$ and
with constant input $v(t)$, satisfies the constraints for all future time instants.
Compared to MPC, the computational load of RG is usually lighter since the cost function does not depend on the state and a single input is computed instead of a longer sequence. Also, RG can guarantee the strong recursive feasibility and constraint satisfaction without additional effort to maintain stability. 
In multirotor control problems, RG has been used to handle obstacle avoidance 
\cite{hermand2018constrained}
and attitude constraints \cite{liu2022constraint}. As far as the authors' knowledge, this is the first work that uses RG for perception-driven constrained control of multirotors.

In order to effectively handle the nonlinear, non-convex visibility constraint set, we approximate it in polynomial form and we tighten it to be contained in the original constraint set. RG is designed with the tightened polynomial constraints by generalizing the work \cite{burlion2022reference}. Unlike \cite{burlion2022reference}, the proposed RG can handle online time-varying references, as required for general trajectory tracking control problems.

The contributions of our work can be summarized as:
\begin{itemize}
    \item We propose a constrained control scheme based on RG able to handle visibility constraints and arbitrary spatial reference inputs for multirotors with guaranteed recursive feasibility and constraint satisfaction. 
    \item We extend the RG for linear systems with polynomial constraints recently introduced in \cite{burlion2022reference} to deal with arbitrary reference inputs. This requires to design suitable constraints on the steady state of the system. The properties of the solution are theoretically proved.
    \item We validated the proposed method with multiple simulations and an hardware experiment. 
\end{itemize}

The paper is organized as follows: 
Sec. \ref{sec_dyn_control} provides an explanation of the dynamics and control approach for the multirotor. Sec. \ref{sec_vis_const} presents a description of the visibility constrained problem, while Sec. \ref{sec_gen_ref_gov} details the proposed RG with its theoretical properties. The proposed method is validated through simulations and hardware experiment in Sec. \ref{sec_validation}. Finally, the paper concludes with Sec. \ref{sec_conclusion}.

\subsection{Notations}
\begin{figure}[t]
    \centering
    \includegraphics[width=0.8\linewidth]{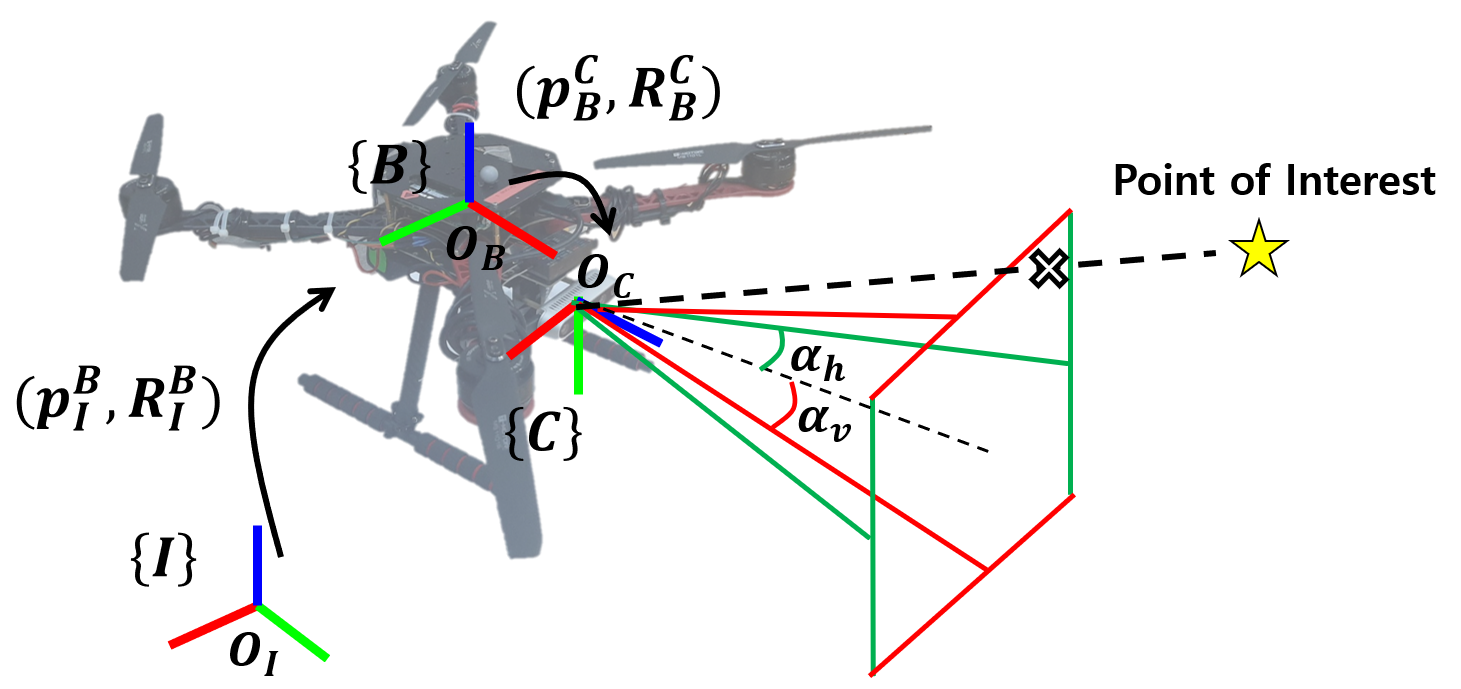}
    \caption{Diagram of the camera-equipped multirotor with depiction of the frames, the camera Field of View, and the point of interest. }
    \label{fig:scheme}
\end{figure}

$\R$ are the real numbers. We denote by $0_{*}$ and $I_{*}$ the zero and identity matrices of dimension $*$. Let $\mathbf{e}_{3}$ denote the standard unit vector of $z$ axis and let $\times$ denote the cross product. The abbreviations $s(\cdot)$, $c(\cdot)$, and $t(\cdot)$ stands for the functions $\sin(\cdot)$, $\cos(\cdot)$, and $\tan(\cdot)$, respectively.
Given an arbitrary vector $\v \in \mathbb{R}^n$, we denote $\v=[v_1,v_2,\dots,v_n]^T$ and $\v[i]=[v_i,\dots,v_n]^T$.
Denote $\sigma(n,r)=\frac{(n+r-1)!}{(n-1)! \, r!}$ and $\Sigma(n,r)={\sum_{i=1}^{r} \sigma(n,i)}$. A set $\mathcal{O}$ is positively invariant if $x(t_0)\in\mathcal{O}$ then $x(t)\in\mathcal{O}$ $\forall t>t_0$.
The inertial frame, the body frame, and the camera frame, depicted in Fig. \ref{fig:scheme}, are denoted as $\mathcal{I}$, $\mathcal{B}$, and $\mathcal{C}$, respectively. The position of the origin of the body frame $O_\mathcal{B}$ with respect to the inertial frame is denoted as $\p_{I}^{B}$, and the rotation from $\mathcal{I}$ to $\mathcal{B}$ is denoted as $R_{I}^{B}$. Transformations between different frame pairs are defined similarly. Note that the transformation from the inertial frame to the body frame is usually omitted for brevity of notations. The angles $\alpha_h$, $\alpha_v$ are half of the horizontal and vertical angles of camera FoV.  

\section{Multirotor Dynamics $\&$ Control} \label{sec_dyn_control}

\subsection{Multirotor Dynamics}
The multirotor dynamics can be expressed as 
\begin{align}
    \dot{\p} &= \textbf{\textit{v}}, \label{pos_dot} \\
    \dot{\textbf{\textit{v}}} &= -g \mathbf{e}_{3} + 
    f R \mathbf{e}_{3},
     \\
     \dot{\boldsymbol{\xi}} &= S \mathbf{\Omega},
    \\
    \dot{\mathbf{\Omega}} &= - J^{-1} (\mathbf{\Omega} \times J\mathbf{\Omega}) + J^{-1}\boldsymbol{\tau},
\end{align}
where $\p\in \R^{3}$ is the position of the multirotor with respect to the inertial frame, $\textbf{\textit{v}} \in \R^{3}$ is the velocity, $\boldsymbol{\xi}=[\varphi, \theta, \psi]^{T}$ is the roll-pitch-yaw Euler angles of the multirotor, $\mathbf{\Omega} \in \R^{3}$ is the angular velocity with respect to the body frame, $S$ is the transformation matrix of the attitude rate $\dot{\boldsymbol{\xi}}$ and the angular velocity $\mathbf{\Omega}$, $J\in \R^{3\times 3}$ is the moment of inertia matrix, $f\in \R$ is the mass-normalized thrust, and $\boldsymbol{\tau}\in \R^{3}$ is the torque vector.

\subsection{Multirotor Control}

\begin{figure}[t]
    \centering
    \includegraphics[width=0.98\linewidth]{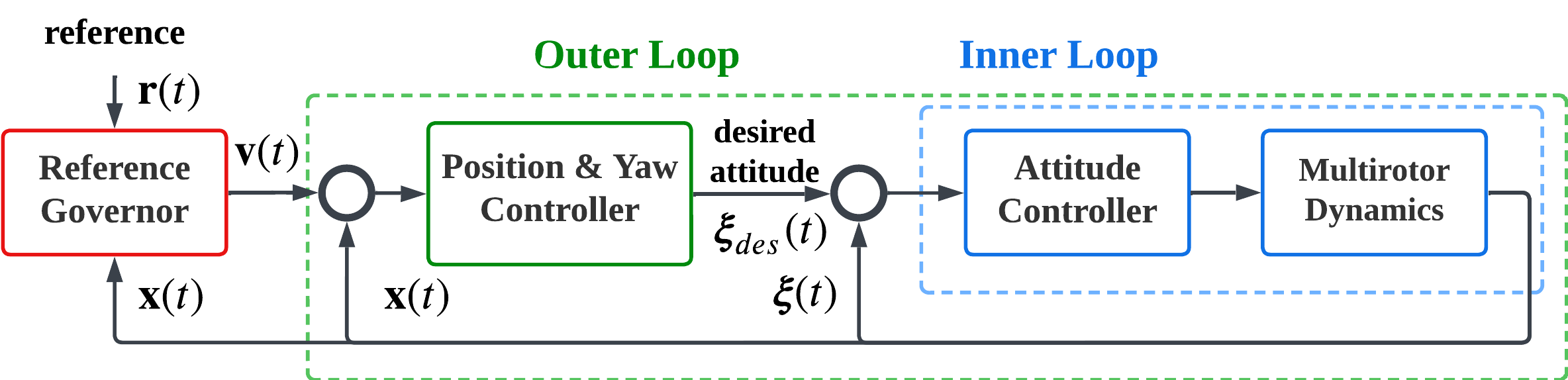}
    \vspace{-2pt}
    \caption{Cascaded control of the multirotor. The RG modifies the given reference $\r(t)$ to convert it to the admissible reference $\v(t)$. The outer-loop controller computes the desired attitude $\boldsymbol{\xi}_{des}(t)$ and the inner-loop controller computes the low-level input to track the desired attitude.}
    \label{fig:control}
\end{figure}

The multirotor dynamics satisfies the \textit{differential flatness} property \cite{mellinger2011minimum}, namely there exists an output, called flat output, such that all the states and inputs of the system can be expressed as function of it and its derivatives. An example of flat output of the multirotor dynamics is $\mathbf{f}_{o} = [\p^T,\psi]^{T}$.

Utilizing the differential flatness property, we use a cascaded control scheme that separates the control problem into outer and inner loops as in Fig. \ref{fig:control}. The outer loop controls the flat output and computes the attitude reference, the inner loop tracks the given attitude reference and generates the low-level input to the system. 

We assume that the inner-loop high-bandwidth attitude controller has negligible regulation delay and local asymptotic stability. The assumption is valid with popular attitude controllers for multirotors such as \cite{lee2010geometric}. For the outer-loop controller with reference $\v$, we use a stabilizing feedback controller (i.e. nonlinear feedback linearization). The resulting closed-loop system is
\begin{align}
    \dot{\x}
&= A_{c} \x + B_{c} \v, \label{sys_dyn}
\end{align} \hspace{-2pt}
where $\x=[\p^T, \psi, \textbf{\textit{v}}^T, \dot{\psi}]^{T}\in \R^{8}$, $\v \in \R^{4}$ is the position and yaw reference vector, which corresponds to the flat output, and $A_c, B_{c}$ are the closed-loop continuous-time system matrices.

\section{Visibility Constrained Control Problem} \label{sec_vis_const}
\subsection{Constraint Formulations}
For safe vision-based tasks, the multirotor is required to keep a PoI in the FoV of the camera.
We formalize this requirement as a constraint on the multirotor state.

Let $\p_{I}^{l}=[x_{I}^{l},y_{I}^{l},z_{I}^{l}]^T$ be the position of the PoI in the inertial frame. As in \cite{jacquet2020motor}, the position of the PoI in the camera frame $\p_{C}^{l}=[x_{C}^{l},y_{C}^{l},z_{C}^{l}]^T$ can be computed with the pinhole camera model as
\begin{align}
    \p_{C}^{l} = R_{B}^{C,T}(R^{T}(\p_{I}^{l}-\p)-\p_{B}^{C}).\label{p_cl}
\end{align}

The visibility constraint can be obtained by requiring that the PoI is inside the limits of the FoV and that the PoI is in front of the camera.
This can be expressed with the inequalities
\begin{subequations} \label{vis_const_total}
\begin{align}
    \Big\lvert x_{C}^{l}/z_{C}^{l} \Big\rvert &\leq t(\alpha_{h}), \label{vis_const_h} \\
    \Big\lvert y_{C}^{l}/z_{C}^{l} \Big\rvert &\leq t(\alpha_{v}), \label{vis_const_v}\\
    z_{C}^{l} &> 0, \label{vis_const_z}
\end{align}
\end{subequations}
where $\alpha_h$ and $\alpha_v$ are half of the horizontal and vertical angles of the camera FoV as in Fig. \ref{fig:scheme}. We call \eqref{vis_const_h}, \eqref{vis_const_v} as the \textit{bearing constraints} and \eqref{vis_const_z} as the \textit{distance constraint}.

In addition to the visibility constraints, in order to take into account the limited capabilities of the multirotor hardware, we require that the linear velocity and acceleration satisfy
$\|\dot{\p}\|_{\infty} \leq \textit{v}_{max}, \|\dot{\textbf{\textit{v}}}\|_{\infty}\leq a_{max}$, where $\textit{v}_{max}$ and $a_{max}$ are the velocity and acceleration limits.

\subsection{Approximations of the Visibility Constraints}
Since roll and pitch angles are usually small and they affect visibility only marginally, we use a simplified form of the constraints to focus on the dependence on position $\p$ and yaw angle $\psi$, as commonly done in the literature \cite{mellinger2011minimum}. To this end, inspired by the visual servoing literature \cite{zheng2019toward}, we introduce the virtual camera frame. It is defined with same position and orientation of the camera frame but roll and pitch angles of the multirotor are set to zero. 
Thus, the rotation of the virtual camera frame with respect to the inertial frame is $R_{I}^{V}=R_{Z}(\psi)R_{B}^{C}$ and the coordinates $\p_{V}^{l}=[x_{V}^{l},y_{V}^{l},z_{V}^{l}]^T$ of the PoI in the virtual frame are
\begin{align}
    R_{B}^{C}\p_{V}^{l}+\p_{B}^{C} &= R_{Z}^{T}(\psi)(\p_{I}^{l}-\p_{I}^{B}),
    \label{virtual_point}
\end{align}
where $R_{Z}(\cdot)$ is the rotation with respect to the $z$ axis.
For convenience of presentation, we assume that the camera is forward-facing and attached to the center of the multirotor, namely $\p_{B}^{C}=0_{3\times1}$.
It follows that
\begin{subequations}    \label{virtual_full_pos}
\begin{align}
    x_{V}^{l} &= s (\psi) (x_{I}^{l} - x_{I}^{B})- c (\psi) (y_{I}^{l}-y_{I}^{B} ), \\
    y_{V}^{l} &= -(z_{I}^{l} - z_{I}^{B}), \\
    z_{V}^{l} &= c (\psi) (x_{I}^{l}-x_{I}^{B}) + s (\psi) (y_{I}^{l} - y_{I}^{B}). 
\end{align}
\end{subequations}

Due to the trigonometric functions of $\psi$, the visibility constraints result to be nonlinear. We propose to approximate the trigonometric functions with polynomial functions. 
For the domain $\psi \in[-\pi/2,\pi/2]$, using \textit{Remez exchange algorithm} \cite{fraser1965survey}, we can find a set of coefficients $\{k_{s}^{i}\}_{i=0}^{n}$, $\{k_{c}^{i}\}_{i=0}^{n}$ so that $s(\psi) \approx f_{s}(\psi)=\sum_{i=0}^{n}k_{s}^{i}\psi^{2i+1}$ and $c(\psi) \approx f_{c}(\psi)=\sum_{i=0}^{n}k_{c}^{i}\psi^{2i}$. By substituting the approximated $\sin$ and $\cos$ functions in \eqref{virtual_full_pos}, the visibility requirement is approximated as polynomial constraints.

The resulting polynomial constraint set is not an inner approximation of the original constraint set. However, the following Lemma shows that the violations of the original constraints for a point satisfying the polynomial constraints are bounded.
\begin{lemma} \label{lemma_bound}
    Assume that $|\varphi| \leq \varphi_{max}<\pi/2$, $|\theta| \leq \theta_{max}<\pi/2$, and $t(\theta_{max})<\min\{t(\alpha_{v})/(1-\Delta_{max}),(1-\Delta_{max})/t(\alpha_{v})\}$
    where $\Delta_{max}=\max |f_{s}^{2}+f_{c}^{2}-1|\ll1$.
    Then, for any point satisfying the approximated constraints, the original distance constraint \eqref{vis_const_z} is satisfied and violations of the bearing constraints \eqref{vis_const_h}-\eqref{vis_const_v} are bounded.
\end{lemma}
\begin{proof}
    The proof is provided in Appendix A.
\end{proof}
We can tighten the approximated constraints by the violation bound of the previous Lemma so that enforcing the resulting constraint set guarantees the satisfaction of the true visibility constraints.
It is possible to numerically show that the approximation error is sufficiently small in practical multirotor settings.
A more detailed explanation is provided in the Appendix.

\subsection{Visibility Constrained Control Problem and Algorithm}
Based on the constraints introduced in Sec. \ref{sec_vis_const}, the problem of visibility constrained control is stated as below. 
\begin{problem}
Find an admissible input trajectory $\v(t)$ such that the state trajectory $\x(t)$ satisfies the visibility constraints \eqref{vis_const_total} with user-defined state constraints and tracks as close as possible the desired reference trajectory $\r(t)$.
\end{problem}

The visibility constrained control problem is solved using RG, which modifies the desired reference command in order to enforce the pointwise-in-time constraints. 
In this work, we propose to extend the RG for linear systems with polynomial constraints \cite{burlion2022reference} in order to tackle arbitrary time-varying references. Further details and the theoretical properties of the proposed algorithm are provided in the following section. 

\section{Generalized Reference Governor for Polynomial Constraints} \label{sec_gen_ref_gov}


\subsection{Background and problem formulation}
Consider a linear discrete-time system
\begin{equation}\label{eq:sysRG}
	\x(k+1) = A\x(k) + B\v(k),
\end{equation}
with $\x \in \mathbb{R}^n$ and $\v \in \mathbb{R}^m$. We assume that the matrix $A$ is stable. 
We introduce the extended system 
\begin{equation}
	\z(k+1) = 
	\left[\begin{array}{cc}
		A & B \\ 0 & I_m
	\end{array}\right] \z(k)
	= \phi \z(k),
\end{equation}
with $\z=[\x^T,\v^T]^T\in \mathbb{R}^{n+m}$. 
Consider the set of constraints
\begin{equation}
	c_i(\z) \leq 0, \qquad i=1,2,\dots,c
\end{equation}
where $c_i:\mathbb{R}^{n+m} \rightarrow \mathbb{R}$ is a polynomial function with monomials up to degree $p$.
Following \cite{kolmanovsky1998theory}, the maximal output admissible set (MOAS) of system \eqref{eq:sysRG} is defined as
\begin{equation}
	O_\infty = \{ \z : c_i(\phi^k \z) \leq 0, \  i=1,2,\dots,c, \ \forall k \geq0\},
\end{equation}
which can be used to enforce the constraints on the evolution of the system.
Unfortunately, in general, $O_\infty$ consists of infinitely many inequalities and thus cannot be computed.
In the case of linear inequalities, there exists an inner approximation of $O_\infty$ that is \textit{finitely determined}, i.e. it can be computed with an iterative procedure in a finite number of steps \cite{kolmanovsky1998theory}. In the case of polynomial constraints, similar procedures to compute an inner approximation exist but they are usually conservative \cite{bemporad1998reference}, rely on Sum of Squares Programming \cite{cotorruelocomputation}, or require $\phi$ to be stable \cite{burlion2022reference}. In this work, we propose to extend the approach of \cite{burlion2022reference} for the case of $\phi$ marginally stable. This allows us to tackle also non-zero time-varying reference inputs.

\subsection{Construction of Maximal Output Admissible Set} \label{sec:moas}
Preliminarily, we define recursively
\begin{equation}
	\z^\expc{r+1} \!=\!
	\left[\!
		z_1 \z[1]^\expct{r} \  z_2 \z[2]^\expct{r} \ \cdots  \ z_n \z[n]^\expct{r} 
	\! \right] ^T
\end{equation}
starting from $\z^\expc{1}=\z$. Note that $\z^\expc{r} \in \mathbb{R}^{\sigma(n,r)}$ is a vector containing all the monomials of degree $r$ obtained from the entries of $\z$ \textit{without} repetitions.
Similarly, we define recursively
\begin{equation}
	\z^\expe{r} = \z \otimes \z^\expe{(r-1)}
\end{equation}
starting from $\z^\expe{1}=\z$. Note that $\z^\expe{r} \in \mathbb{R}^{(m+n)^r}$ is a vector containing all the monomials of degree $r$ obtained from the entries of $\z$ but \textit{with} repetitions.
There always exist two matrices $M_c(n\!+\!m,r)$ and $M_e(n\!+\!m,r)$ such that
\begin{equation}\label{eq:McMe}
	\z^\expc{r} = M_c(n+m,r) \z^\expe{r}, \quad \z^\expe{r} = M_e(n+m,r) \z^\expc{r}.
\end{equation}
A procedure to compute $M_c(n+m,r)$ and $M_c(n+m,r)$ can be obtained following \cite{valmorbida2013design}. 
By recursively defining
\begin{equation}
	\phi^\expe{r}=\phi \otimes \phi^\expe{(r-1)}
\end{equation}
starting from $\phi^\expe{1}=\phi$, we have that
\begin{equation}\label{eq:sys_poly_e}
	\z^\expe{r}(k+1) = \left(\phi \z(k) \right)^\expe{r} = \phi^\expe{r} \z^\expe{r}(k),
\end{equation}
where we used the mixed-product property of the Kronecker product. Using \eqref{eq:McMe} and \eqref{eq:sys_poly_e}, we obtain
\begin{align}
	\z^\expc{r}(k+1) 
	&= M_c(n+m,r) \z^\expe{r}(k+1)  \\
	&= M_c(n+m,r) \phi^\expe{r} \z^\expe{r}(k)  \\
	&= M_c(n\!+\!m,r) \phi^\expe{r}  M_e(n\!+\!m,r) \z^\expc{r}(k) \\
	&=\phi^\expc{r} \z^\expc{r}(k). \label{eq:sys_poly_c}
\end{align}
The following Lemma provides the spectral characterization of $\phi^\expc{r}$ and $\phi^\expe{r}$.
\begin{lemma}\hspace{-4pt} \label{lemma_phi}
	The matrix $\phi^\expe{r}$ is marginally stable and has exactly $m^r$ unitary eigenvalues. 
	The matrix $\phi^\expc{r}$ is marginally stable and has exactly $\sigma(m,r)$ unitary eigenvalues. 
\end{lemma}
\begin{proof}
     Proof is given in Appendix B.
\end{proof}

We can now stack the vectors $\z^\expc{r}$ for $r=1,\dots,p$ as
\begin{equation}
	\overline{\ZZ} = \left[ \z^\expct{1} \ \z^\expct{2} \ \cdots \ \z^\expct{p} \right]^T 
\end{equation}
obtaining a vector containing all the monomials up to degree $p$ obtained from entries of $\z$ without repetitions. Moreover, we can introduce the auxiliary linear system 
\begin{equation}
	\overline{\ZZ}(k\!+\!1) = 
	\left[\begin{smallmatrix}
		\phi^\expc{1} & & & 0 \\ & \phi^\expc{2} & & \\ & & \dots & \\ 0 & & & \phi^\expc{p} 
	\end{smallmatrix}\right] 
	\overline{\ZZ}(k)
	=\overline{\Phi}\hspace{1pt} \overline{\ZZ}(k).
\end{equation}
Exploiting $\overline{\ZZ}$ the polynomial inequalities $c_i(\z)\leq 0 $ can be equivalently expressed by linear constraints $C_i \overline{\ZZ} \leq c_{i0}$. 
In this way, as done in \cite{burlion2022reference}, it is possible to exploit the theory of MOAS for linear systems with linear constraints to compute an approximation of the MOAS of the original linear system with polynomial constraints. However, the solution proposed in \cite{burlion2022reference} requires that $\phi$ is stable and it can be used only with null reference. In this work, instead of using the vector $\overline{\ZZ}$, in order to easily find an inner approximation of the MOAS that can be used for any reference input, we permute the components of the state to highlight the subsystem containing unitary eigenvalues. With a little abuse of notation, denote 
\begin{equation}
	\z^\expc{r+1}_x \!=\! \left[  x_1 \z[1]^\expct{r} \, x_2 \z[2]^\expct{r} \cdots \, x_n \z[n]^{\expct{r}}  \right]^T.
\end{equation}
Note that $\z^\expc{r}=[\z^\expct{r}_x, \v^\expct{r}]^T$.
Now we can define
\begin{align}
	\ZZ_x  \,   &= \left[ \z^\expct{1}_x \ \z^\expct{2}_x \ \cdots \ \z^\expct{p}_x \right]^T \\
	\VV    \ \, &= \left[ \v^\expct{1}   \ \v^\expct{2}   \ \cdots \ \v^\expct{p} \right]^T
\end{align}
and introduce $\ZZ \in \mathbb{R}^{\Sigma(n,p)}$ as
\begin{equation}
	\ZZ = \eta(\z) =
	\left[\begin{array}{c c}
		\ZZ_x^T & \VV^T
	\end{array}\right]^T,
\end{equation}
where $\eta(\cdot)$ is a suitable function. Since $\ZZ$ is a permutation of the vector $\overline{\ZZ}$, there always exists a matrix $T$ such that
\begin{equation}
\ZZ = T \overline{\ZZ}
\end{equation}
and
\begin{equation}\label{eq:sys_poly_stack}
	\ZZ(k+1)= \Phi \ZZ(k) = T \overline{\Phi} T^T \ZZ(k)\;.
\end{equation}

\begin{lemma} \label{lemma:Large_phi}
	It holds that
	\begin{equation}
		\Phi =
		\left[\begin{array}{cc}
			F & G \\ 0 & I_{\Sigma(m,p)}
		\end{array}\right],
	\end{equation}
	where $F$ is stable.
\end{lemma}
\begin{proof}
	The proof follows from the Lemma \ref{lemma_phi} since $\Phi$ is a permutation of the block diagonal matrix with diagonal blocks $\phi^\expc{r}$, $r=1,\dots,p$. 
\end{proof}
It is possible to rewrite the polynomial function $c_i$ as
\begin{equation}
	c_i(\z)=\sum\nolimits_{r=1}^p c_{ir} \z^\expc{r} - c_{io} = C_i \ZZ - c_{io}\;.
\end{equation}
We introduce the following set
\begin{equation}
	\Z = \{ \ZZ : C_i \ZZ \leq c_{i0}, \ i=1,2,\dots,c \} .
\end{equation}
The maximal output admissible set of \eqref{eq:sys_poly_stack} is defined as
\begin{equation}
	O'_\infty = \{ \ZZ : \Phi^k \ZZ \in \Z, \  \forall k \geq0\} \;.
\end{equation}
Based on the set
\begin{equation}
	O'_\epsilon = \{ \ZZ=(\ZZ_x, \VV) : ((I-F)^{-1} G \VV,\VV) \in (1-\epsilon)\Z \}
\end{equation}
for $\epsilon>0$, we can define the following inner approximation
\begin{equation*}
	\tilde{O}'_\infty = O'_\infty \cap O'_\epsilon.
\end{equation*}

\begin{proposition}\label{th:moas}
	If $\Z$ is compact,  $\tilde{O}'_\infty$ is \textit{finitely determined}, and \textit{positively invariant}.  
\end{proposition}
\begin{proof}
	From Lemma \ref{lemma:Large_phi}, $\Phi$ is marginally stable. Then, the statement follows from Theorem 7.2 in \cite{kolmanovsky1998theory} by rewriting the system \eqref{eq:sys_poly_stack} in the required form. See also in \cite[Page 355]{kolmanovsky1998theory}. 
\end{proof}
As outlined in \cite{kolmanovsky1998theory}, the tighter condition on the steady-state enforced by $O'_\epsilon$ is fundamental in order to guarantee that $\tilde{O}'_\infty$ is finitely determined. 
Now we can define 
\begin{equation}\label{eq:moas}
	\tilde{O}_\infty = \{ \z : \ZZ \in \tilde{O}'_\infty, \ZZ=\eta(\z) \}.
\end{equation}
Then we have the following proposition.
\begin{proposition} \label{prop2}
	If $\Z$ is compact,  $\tilde{O}_\infty$ is \textit{finitely determined}, \textit{positively invariant}, and $\tilde{O}_\infty \subseteq O_\infty$
\end{proposition}
\begin{proof}
	The statement follows from the previous Proposition and the fact that, if $\ZZ(k)=\eta(\z(k))$, then $\ZZ(k+1)=\eta(\z(k+1))$ since $\eta(\phi \z) = \Phi \eta(\z)$.
\end{proof}

Since the set $\tilde{O}'_\infty$ is the MOAS for a linear system subject to linear constraints, it can be computed following the well-known iterative procedure proposed in  \cite{kolmanovsky1998theory}. According to \eqref{eq:moas}, $\tilde{O}_\infty$ can be immaterially obtained from $\tilde{O}'_\infty$. Note that $\tilde{O}_\infty$ is an inner approximation of $O_\infty$ and it can be taken arbitrarily close to $O_\infty$ by choosing $\epsilon$ arbitrarily close to $0$.
Differently from \cite{burlion2022reference}, we require $\ZZ \in O'_\epsilon$. In this was, $\phi$ does not need to be stable. 

The assumption on the compactness of $\mathcal{Z}$ can be relaxed by introducing the constraint output $\YY=[C, D]\ZZ$ and the constraint set $\mathcal{Y}=\{ \YY: H_i\YY\leq h_{io}, i=1,\dots,c\}$  with suitable $C,D,H_i,h_{io}$ such that conditions $[C, D]\ZZ\in\mathcal{Y}$ and $\ZZ\in\mathcal{Z}$ are equivalent. In this case, Proposition \ref{th:moas} holds if $(\Phi,C)$ is observable and $\mathcal{Y}$ is compact. Fictitious redundant constraints can be added to meet this requirement following the procedure from Remark 5 of \cite{burlion2022reference}.

\subsection{Reference Governor}
The RG can be formalized as
\begin{align}
	\lambda(k) &= \arg\max\nolimits_{\lambda\in[0,1]} \, \lambda \\
	 &\text{s.t. } (\x(k), \v(k\!-\!1) \!+\! \lambda(\r(k)\!-\!\v(k\!-\!1))) \in \tilde{O}_\infty
\end{align}
and $\v(k)=\v(k\!-\!1) + \lambda(k)(\r(k)-\v(k\!-\!1))$.
The above optimization problem can be solved by bisection. 

Since the MOAS is not limited to the case with $\phi$ stable, the proposed method can be used for any arbitrary time-varying reference $\r(k)$. This generalizes the solution proposed by \cite{burlion2022reference}, which can be used only to drive the system to the origin, i.e. $r(k)=0$. 
Although \cite{burlion2022reference} can be applied for non-null references by computing the MOAS for the error dynamics, this approach is impractical to handle time-varying references since it requires re-computing the MOAS at every time step. 
Furthermore, even for a fixed reference, the set of feasible initial conditions is enlarged with the proposed strategy. Roughly speaking, this is due to the fact that \cite{burlion2022reference} requires that the initial state admits the existence of a sequence of inputs linearly converging to 0 whose corresponding evolution is admissible, 
while the proposed strategy simply requires the existence of an initial admissible input, as stated in the following proposition. 
\begin{proposition}\label{th:rg}
	Assume that there exist $\v_0$ such that  $(\x(0),\v_0) \in \tilde{O}_\infty$, i.e. the optimization problem is feasible at time $k=0$. Then, the optimization problem is recursively feasible and the resulting system evolution satisfies the constraints. 
\end{proposition}
\begin{proof}
Recursive feasibility holds since input $\lambda(k)=0$ is always admissible since MOAS is positively invariant.
Constraints are satisfied because $(x(k),v(k)) \in \tilde{O}_\infty$ implies  that $c_i(x(k),v(k))\leq 0$.
\end{proof}
In general, it is not possible to guarantee the convergence to the global optimal point since constraints are not convex. In that case, convergence can be obtained by using an inner convex approximation of $O'_\epsilon$.  

\subsection{Application to Visibility Constrained Control}
In this subsection, we explain how the previously designed RG can be applied to the visibility constrained control problem. Since deriving the MOAS is computationally demanding, we construct it offline. It is constructed following Sec. \ref{sec:moas} with the polynomial approximation of the visibility constraints and the discrete-time version of the system.\footnote{Note that in order to meet the requirement of compact feasible set in Proposition \ref{prop2}, small positive value $\epsilon_{z}$ is introduced to the distance constraint \eqref{vis_const_z} as  $z_{C}^{l}\geq \epsilon_{z}>0$.} However, since the constraints depend on $\p_{I}^{l}$, the MOAS would depend on it as well and we would need to compute a different MOAS whenever the PoI changes.
We can avoid this by introducing a new frame $\mathcal{L}$ which has the origin at $\p_{I}^{l}$ and the same orientation of the inertial frame $\mathcal{I}$. The MOAS is constructed with the transformed constraints. In this way, when the PoI changes, we do not need to recompute the MOAS but we have to express any positions with respect to the new frame, that can be easily done by a change of coordinate. Using this approach, RG can be used online without prior information of the landmark. 

With the constructed MOAS, the RG is used to modify a given reference trajectory to ensure visibility constraint. The online control method is described in Algorithm \ref{alg: pa_traj_opt}. With the given MOAS $\tilde{O}_{\infty}$, PoI $\p_{W}^{l}$, reference $\r$, and initial state $\x_0$, at each time instant, the state and the desired reference are first transformed to the landmark frame $\L$. Then, RG is used to compute the modified reference in frame $\L$. Subsequently, it is transformed to the inertial frame, and applied to the closed-loop multirotor system \eqref{sys_dyn}.
From Proposition \ref{th:rg}, the proposed method guarantees recursive feasibility and visibility constraint satisfaction. 

\begin{algorithm}[t]
\caption{Visibility Constrained Control with RG} \label{alg: pa_traj_opt}
\begin{algorithmic}[1]
\STATE \textbf{Input}: MOAS $\tilde{O}_{\infty}$, PoI $\p_{I}^{l}$, reference $\{\r\}_{k=1}^{K}$, initial state $\x_{0}$
\FOR{ $k=1,\cdots, K$}
\STATE $\x_{L}$, $\r_{L}\leftarrow$ CoordTransform$_{\mathcal{I}\rightarrow\L} (\x_k$, $\r_k$, $\p_{I}^{l})$
\STATE $\v_{L}\leftarrow$ ReferenceGovernor$(\tilde{O}_\infty,\x_{L},\r_{L},\v_{L,prev})$
\STATE $\v_{L,prev}\leftarrow \v_{L}$
\STATE $\v_k\leftarrow$ CoordTransform $_{\mathcal{L}\rightarrow\mathcal{I}} (\v_{L}, \p_{I}^{l})$
\STATE $\x_{k+1} \leftarrow $ MultirotorDynamics$(\x_k, \v_k)$
\ENDFOR
\end{algorithmic}
\end{algorithm}

\section{Validation} \label{sec_validation}
We validate the proposed RG for visibility constrained control with simulations and a real-world experiment. 

\begin{figure}[t]
    \centering
      \subfloat[]      
{\includegraphics[width=0.5\linewidth]{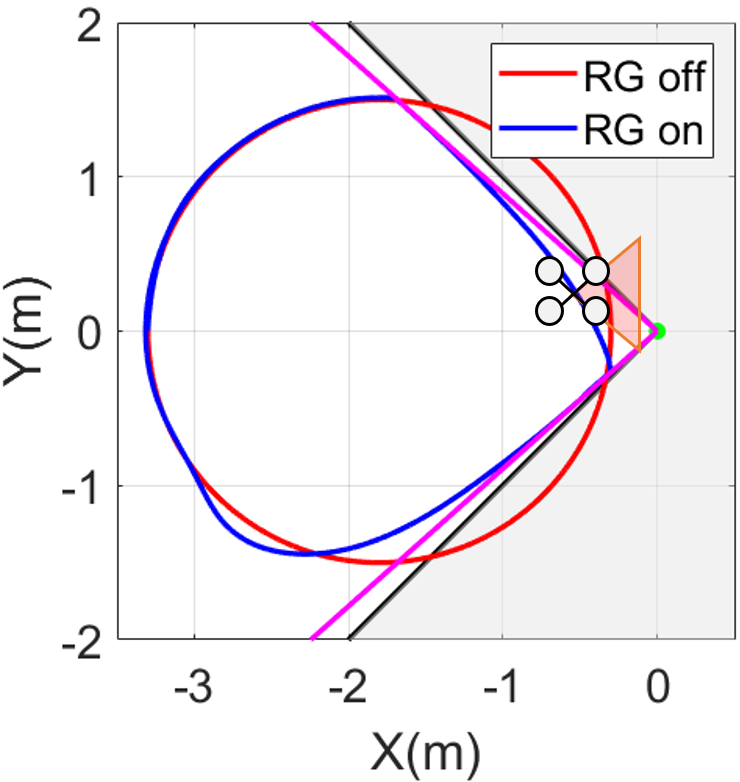}} 
\\
  \subfloat[]      
  {\includegraphics[width=0.75\linewidth]{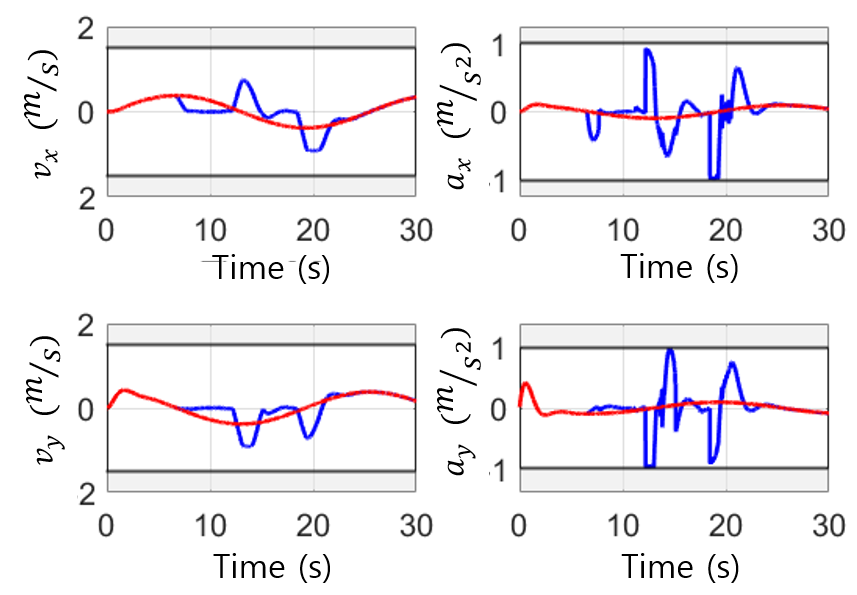}}
\caption{Simulation result of the single PoI scenario for the solution without the proposed RG (red line) and with the proposed RG (blue line). (a) Trajectory of the multirotor with the PoI (green dot) and the non-admissible region (grey area). Boundaries of the admissible region are depicted for the true constraints (black line) and the tightened constraints taking into account the approximation error (magenta line).  (b) Velocity (left) and acceleration (right) of the multirotor.}
\label{fig:result_scenario1}
\end{figure}

\begin{table}[t]
    \captionsetup{justification=centering}
    \caption{Parameter Set for the Simulations}
    \centering
    \begin{tabular}{|c|c|}
    \hline
       Types & Parameter Values \\ \hhline{|=|=|} 
       Discretization Time & $Ts=10\,$ms \\ \hline
       Camera FoV &$\alpha_h =45\degree$, $\alpha_v = 35\degree$ \\ \hline
       \multirow{2}{*}{Polynomial Approx.} & $k_{c}^{0}=0.8798$, $k_{c}^{2}=-0.3566$ \\ 
       & $k_{s}^{1}=0.9928$, $k_{s}^{3}=-0.1462$ \\ \hline
        Max. Roll $\&$ Pitch & $\theta_{max} = 4\degree$, $\varphi_{max} = 4\degree$ \\ \hline
       State Constraints &$\textit{v}_{max}=1.5\,$m/s,  $a_{max}=1.0\,$m/s$^2$ \\ \hline
       Control Gains & $k_P=4$, $k_D=2$
       \\\hline
       \end{tabular}
    \label{tab:parameters}
\end{table}

\subsection{Simulation Setup}
In the following simulations, we assume that the multirotor is stabilized using the control law proposed in \cite{lee2010geometric}. The input force is obtained by proportional-derivative (PD) control with the P gain $k_{P}$ and D gain $k_{D}$. The desired attitude is computed based on the obtained force vector. More details can be found in \cite{lee2010geometric}.
Approximation bounds to tighten the polynomial constraints are computed with maximum roll and pitch angles $\varphi_{max},\theta_{max}$. The system dynamics is discretized with sampling time $T_s$. The simulations are performed in MATLAB environment on a i7 laptop. 
The parameters used in the simulations are listed in Table \ref{tab:parameters}.  

\subsection{Simulation: Single PoI}
In this subsection we consider the case with a single PoI and a circular reference trajectory. The reference trajectory is on a 2D plane with zero yaw angle, $\r(t)=[R c(\omega t+\pi/2)-1.5R, R s(\omega t+\pi/2), 0, 0]^T$ with $R=1.5\,$m, $\omega=2\pi/25$. The PoI is located at the origin. We compare the solution with and without the proposed RG and the results are given in Fig.~\ref{fig:result_scenario1}. The trajectory obtained without the proposed RG can track the reference but loses the visibility of the PoI. Conversely, the trajectory with the proposed RG always satisfies the visibility constraints and other state constraints as shown in Fig. \ref{fig:result_scenario1} (b).
In particular, in the region close to the origin ($x\geq -1$), the system tracks the best admissible approximation of the desired non-admissible reference. As soon as the desired reference is steady-state admissible ($x\leq -1$, $y\leq 0$), the RG drives the system as fast as possible back to the desired reference, causing a small overshoot. It can be probably removed by slowing down the desired reference trajectory.

\begin{figure}[t]
    \centering
    \subfloat[]
    {\includegraphics[width=0.65\linewidth]{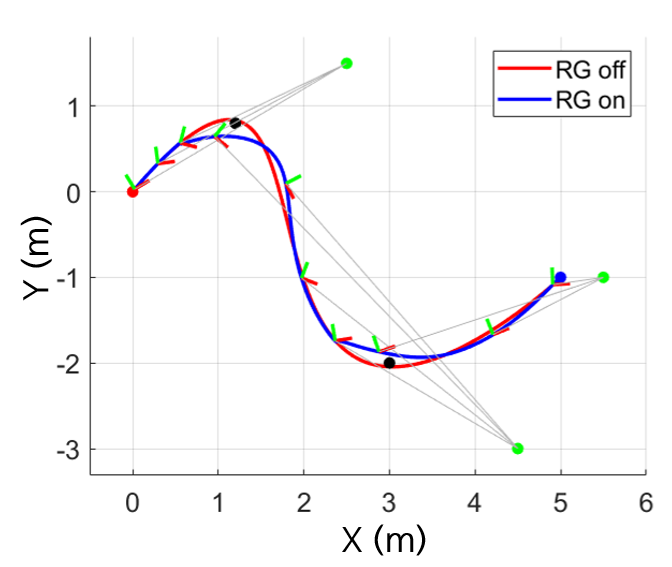}}
    \\
    \subfloat[]
    {\includegraphics[width=0.65\linewidth]{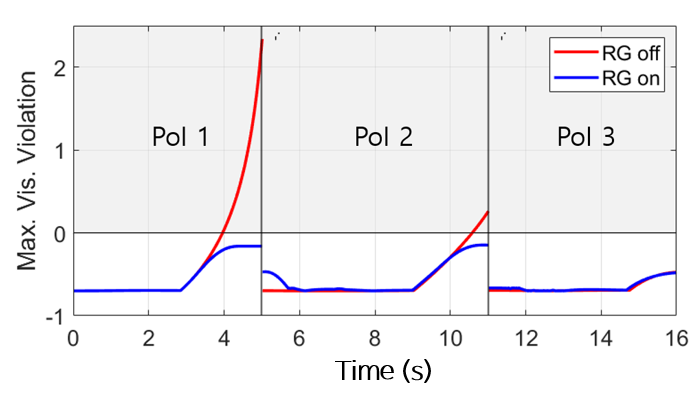}}
    \caption{Simulation result of the multi PoI scenario for the solution without (red line) and with the proposed RG (blue line). (a) Trajectory of the multirotor where the proposed RG follows the reference passing waypoints (black dots) while maintaining the visibility to the PoI (green dots). Multirotor poses (red$\&$green axes) are shown for every 1.5$s$. (b) Maximum violations of the visibility constraints (7a-c).}
    \label{fig:simul_multi_poi}
    \vspace{-4mm}
\end{figure}

\subsection{Simulation: Multiple PoIs}
\vspace{-1mm}
We consider the scenario with multiple PoIs and time-varying yaw reference. From the initial position $s=[0,0,0]^T$ to the goal position $g=[5,-1,0]^T$, the reference trajectory is obtained as the concatenation of three polynomial trajectories with two intermediate waypoints at $w_1=[1.2,0.8,0]^T$ and $w_2=[3,-2,0]^T$. For each piece of the reference trajectory, the multirotor requires to maintain the visibility of a different PoI, namley $\p_I^1=[2.5,1.5,0]^T$, $\p_I^2=[4.5,-3,0]^T$, and $\p_I^3=[5.5,-1,0]^T$.

The result with the proposed RG for this task is given in Fig. \ref{fig:simul_multi_poi}. Fig. \ref{fig:simul_multi_poi}(a) shows how the proposed RG modifies the reference trajectory and Fig. \ref{fig:simul_multi_poi}(b) confirms that the proposed RG is able to prevent violations of the visibility constraints. Thanks to the low computaional demand of RG, since the mean and maximum computation times (equal to $0.89\,$ms and $3.6\,$ms respectively) are smaller than the sampling period, the proposed method can be used for real-time control.


\begin{figure}[t!]
    \centering
    \subfloat[]
    {\includegraphics[width=0.8\linewidth]{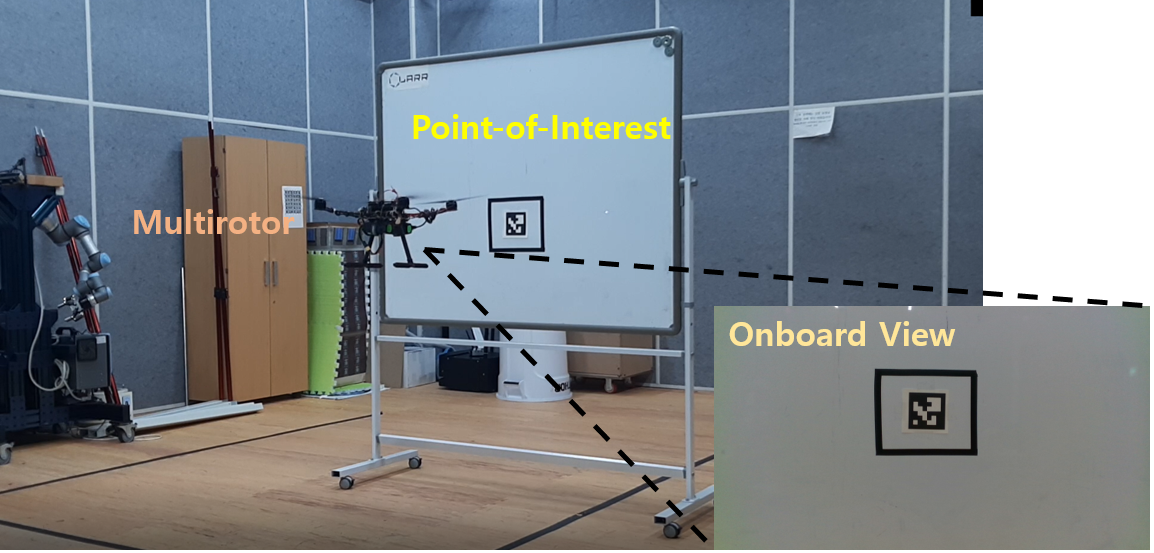}}
    \\
    \subfloat[]
    {\includegraphics[width=0.48\linewidth]{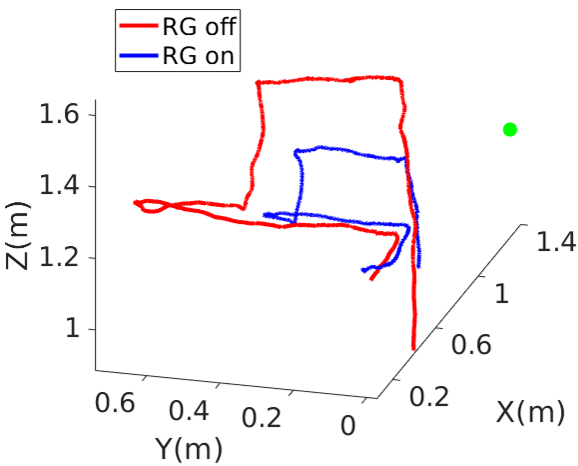}}
    \hspace{1mm}
    \subfloat[]
    {\includegraphics[width=0.42\linewidth]{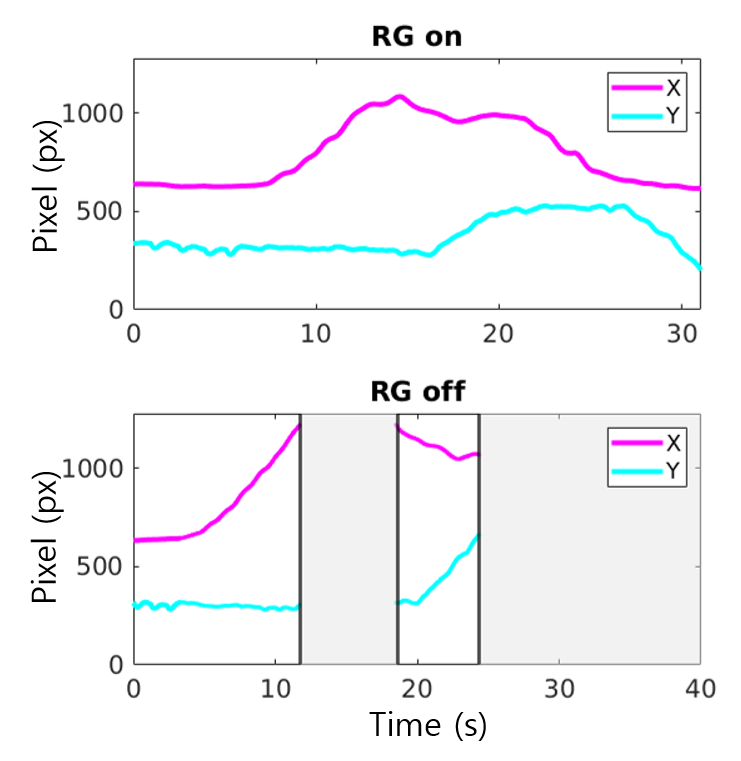}}
    \caption{ (a) Snapshot from the hardware experiment. (b) 3D trajectory of the multirotor and  (c) the image coordinate trajectories of the PoI. While the controller without the proposed RG violates the visibility constraints (grey intervals), the proposed RG can prevent violations during the flight.}
    \vspace{-4mm}
\label{fig:exp}
\end{figure}

\subsection{Hardware Experiment}
In the experiment, we consider the scenario where the multirotor sweeps a wall and  maintain the visibility of a PoI located on the wall. The reference is given by a human pilot, similarly to teleoperations for inspection, and the visibility can be violated due to human errors. 
A quadrotor equipped with a forward-facing Realsense D435 camera ($\alpha_h=34.5\degree$, $\alpha_v=21\degree$) is used. We use the Pixhawk4 flight controller and an Intel i7 NUC computer. The state information is provided by the Optitrack motion capture system. The position of the PoI is assumed to be known.
To identify the model of the closed-loop system, we use the \textit{n4sid} algorithm \cite{van1994n4sid}.

Experimental results are shown in Fig. \ref{fig:exp}. We can see that the proposed RG maintains visibility of the PoI and prevents violations due to wrong commands given by the human pilot. The video of the experiment can be checked at \url{https://youtu.be/SquHiHjRsMQ}.

\section{Conclusion} \label{sec_conclusion}
In this work, we deal with the visibility-constrained control problem for multirotors, where the objective is to track a given reference while maintaining the visibility of a point-of-interest. The visibility constraint is formulated based on the camera geometry and results in nonlinear constraints. The proposed method based on RG guarantees constraint satisfaction and can be used in real-time. Simulations and an hardware experiment validate the proposed method. For future work, we will integrate the proposed method with high-level path planning algorithms to obtain a complete solution for many applications of camera-equipped multirotors. 

\section*{Appendix}
\subsection{Bound of Violation by the Constraint Approximation} \label{append_bound}

\begin{proof}[Lemma 1]
Due to the symmetry of the bearing constraints, we only consider the case when $x_{C}^{l}>0$ and $y_{C}^{l}>0$. The violation functions $g_{1} = x_{C}^{l}/ z_{C}^{l} - t (\alpha_{h})$ and $g_{2} = y_{C}^{l}/z_{C}^{l} - t (\alpha_{v})$ increase with $x_{C}^{l}$ and $y_{C}^{l}$.  
The maximum violation occurs at the boundary of the feasible region, which is $x_{V}^{l}=t(\alpha_{h}) z_{V}^{l}$, $y_{V}^{l}=t(\alpha_{v}) z_{V}^{l}$. 
Let $f_{c}^2(\psi)+f_{s}^2(\psi)-1=\Delta_{\psi}$ where $|\Delta_{\psi}|\leq \Delta_{max}$, and $k = (1+\Delta_{\psi}) c (\theta) - s (\theta) t (\alpha_{v})$. Then, due to the assumption, $k>0$ and $z_{C}^{l} = k z_{V}^{l} >0$, thus the distance constraint is satisfied. 


The violation of the bearing constraints is bounded as \vspace{-1mm}
\begin{align}
    g_{1} &\leq \Big\{ (1+\Delta_{max}) (s (\theta) s (\varphi) + c (\varphi) t (\alpha_h) - c (\theta) t (\alpha_h)) \nonumber \\
    &+ c (\theta) s (\varphi) t (\alpha_{v}) + s (\theta) t (\alpha_v) t (\alpha_h) \Big \}/k_{min} = \epsilon_{1} (\varphi,\theta), \nonumber \\ 
    g_{2} &\leq \Big\{  (1+\Delta_{max}) (s (\theta) c (\varphi) - s (\varphi) t (\alpha_h) - c (\theta) t (\alpha_v)) \nonumber \\
    & + s (\theta) t^2 (\alpha_{v})  + c (\theta) c (\varphi) t (\alpha_{v}) \Big \} /k_{min} = \epsilon_{2}(\varphi,\theta), \nonumber 
\end{align} 
where $k_{min}=(1-\Delta_{max})c(\theta)-s(\theta)t(\alpha_v)$. $k_{min}$ decreases with respect to $\theta\in [-\theta_{max},\theta_{max}]$, and is positive as $k_{min}\geq (1-\Delta_{max})c(\theta_{max})-s(\theta_{max})t(\alpha_{v})>0$ with the assumption.
 $\epsilon_{1}(\varphi,\theta)$ and $\epsilon_{2}(\varphi,\theta)$ are bounded as $\epsilon_{1}(\varphi,\theta)\leq \epsilon_{1,max}$ and $\epsilon_{2}(\varphi,\theta)\leq \epsilon_{2,max}$.
\end{proof} 

\begin{figure}[t]
    \centering
      \subfloat[]      
{\includegraphics[width=0.44\linewidth]{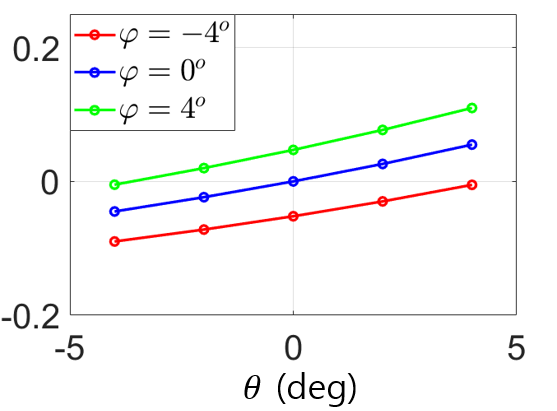}}
  \hspace{2mm}
  \subfloat[]      
  {\includegraphics[width=0.45\linewidth]{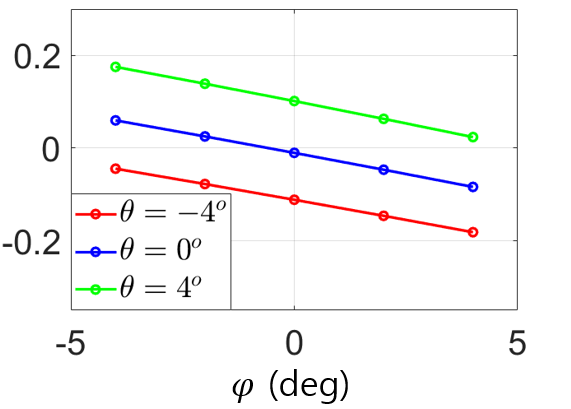}}
\caption{The bound of the violations of the bearing constraints for different $\phi$, $\theta$ computed from (a) $\epsilon_{1}(\phi,\theta)$ and (b) $\epsilon_{2}(\phi,\theta)$. }
\label{fig:error_bound}
  \end{figure}

With the parameters from Table \ref{tab:parameters}, the violation bounds are computed as in Fig. \ref{fig:error_bound}. The maximum bound is used to tighten the visibility constraints. Since the camera FoV is $\alpha_h=45\degree$, $\alpha_v=35\degree$ and the bounds are $\epsilon_{1,max}=0.110$, $\epsilon_{2,max}=0.175$, the reduced FoV is $\alpha_h=41.7\degree$, $\alpha_v=27.7\degree$. It can be used to guarantee the feasibility of the true visibility constraints.

\subsection{Proof of Lemma \ref{lemma_phi}}
We start with useful properties of the Kronecker product.

\begin{lemma}\label{lem:kronecker}
    The following statements hold:
	\begin{enumerate}
        \item $ (A \otimes B) \otimes C = A \otimes (B \otimes C)$
	\item $AB \otimes CD = (A \otimes C)(B \otimes D)$
	\item  Let $\lambda^A_i$, $i\!=\!1,\dots,n$, be the eigenvalues of $A$ and $\lambda^B_j$, $j\!=\!1,\dots,m$, be the eigenvalues of $B$. Then $\lambda^A_i \lambda^B_j$, $i\!=\!1,\dots,n$, $j\!=\!1,\dots,m$, are the eigenvalues of $A\otimes B$.
	\end{enumerate} 
\end{lemma}
We are now ready to prove Lemma \ref{lemma_phi}. 

\begin{proof}
    We prove the first statement by inductive argument. It holds for $r=1$ since $\phi^\expe{1}=\phi$. Assume that the statement holds for $r-1$. Then
	\begin{equation*}
		\phi^\expe{r} = \phi \otimes \phi^\expe{r-1} = 	
		\left[\begin{array}{cc}
			A \otimes \phi^\expe{r-1} & B \otimes \phi^\expe{r-1} \\ 0 \otimes \phi^\expe{r-1} & I_m \otimes \phi^\expe{r-1}
		\end{array}\right].
	\end{equation*}
	Since $A$ is stable and $\phi^\expe{r-1}$ is marginally stable for inductive hypothesis, $A \otimes \phi^\expe{r-1}$ is stable due to Lemma~\ref{lem:kronecker}.3. Since $I_m \otimes \phi^\expe{r-1}$ is the block diagonal matrix containing $m$ copies of $\phi^\expe{r-1}$ and $\phi^\expe{r-1}$ is marginally stable with exactly $m^{r-1}$ unitary eigenvalues by inductive hypothesis, $I_m \otimes \phi^\expe{r-1}$ is marginally stable with $m^{r}$ unitary eigenvalues. \\
    To prove the second statement, let $T_r$ be a change of basis matrix such that $\z_r=T_r \z^\expe{r}$ with $\z_r = [*, \, \v^\expet{r}]^T$ and 
    $\z_r(k+1) = T_r \, \phi^\expe{r} \, T_r^{-1} \z_r(k)$.
	Since $\v^\expe{r}(k+1)=\v^\expe{r}(k)$, it follows that
 $\phi_r = \left[\begin{smallmatrix}
     F_r & G_r  \\ 0 & I_{m^r}
 \end{smallmatrix}\right]$
	for some matrices $F_r$  and $G_r$. 
    Since $\phi^\expe{r}$ is marginally stable with exactly $m^r$ unit eigenvalues and $\phi_r$ is equivalent to $\phi^\expe{r}$, $F_r$ is stable.
	
	Recall that there exist two matrices $M_c(m,r) \in \mathbb{R}^{\sigma(m,r) \times m^r}$ and $M_c(m,r) \in \mathbb{R}^{m^r \times \sigma(m,r)}$ such that $\v^\expc{r} = M_c(m,r) \v^\expe{r}, \v^\expe{r} = M_e(m,r) \v^\expc{r}$.
    Since $M_e(m,r)$ is full-column rank, $M_c(m,r)$ can be computed as the Moore-Penrose pseudo-inverse of $M_e(m,r)$ and $M_c(m,r)M_e(m,r)=I_{\sigma(m,r)}$.
	Similarly, there always exist two matrices $N_c(n\!+\!m,r) \in \mathbb{R}^{\sigma(n\!+\!m,r) \times (n\!+\!m)^r}$ and $N_c(n\!+\!m,r) \in \mathbb{R}^{(n+m)^r \times \sigma(n+m,r)}$ such that 
 $\z^\expc{r} = N_c(n+m,r) \z_r$, $\z_r = N_e(n+m,r) \z^\expc{r}$.
	Since  $\z^\expc{r} = [*\,, \v^\expct{r}]^T$ and $\z_r = [*\,, \v^\expet{r}]^T$, it holds that
	\begin{align*}
		N_c(n+m,r) &=
		\left[\begin{array}{cc}
			N^{11}_c &  N^{12}_c \\ 0 & M_c(m,r) 
		\end{array}\right],  \\
		N_e(n+m,r) &=
		\left[\begin{array}{cc}
			N^{11}_e &  N^{12}_e \\ 0 & M_e(m,r) 
		\end{array}\right].     
	\end{align*} 
	We have
	\begin{align*}
		\z^\expc{r}(k+1) 
		&= N_c(n+m,r) \z_r(k+1)  \\
		&= N_c(n+m,r) \phi_r \z_r(k)  \\
		&= N_c(n+m,r) \phi_r  N_e(n+m,r) \z^\expc{r}(k).
	\end{align*}
	From \eqref{eq:sys_poly_c} we must have
	\begin{align*}
		\phi^\expc{r} 
		&= N_c(n+m,r) \phi_r  N_e(n+m,r) \\
		&=
		\left[\!\!\begin{array}{cc}
			N^{11}_c &  N^{12}_c \\ 0 & M_c(m,r) 
		\end{array}\!\!\right] \!\!
		\left[\!\!\begin{array}{cc}
		F_r & G_r  \\ 0 & I_{m^r} 
		\end{array}\!\!\right] \!\!
		\left[\!\!\begin{array}{cc}
			N^{11}_e &  N^{12}_e \\ 0 & M_e(m,r) 
		\end{array}\!\!\right] \\
		&=
		\left[\begin{array}{cc}
			N^{11}_c F_r N^{11}_e & * \\ 0 &  M_c(m,r)M_e(m,r)
		\end{array}\right].
	\end{align*}
    Following the same procedure we can obtain 
	\begin{equation*}
		\z^\expc{r}(k) = N_c(n+m,r)  \left(\phi_r\right)^k N_e(n+m,r) \z^\expc{r}(0), 
    \end{equation*}
    \begin{equation*}
        \left(\phi^\expc{r}\right)^k =
		\left[\begin{array}{cc}
			N^{11}_c \left(F_r\right)^k N^{11}_e & * \\ 0 &  M_c(m,r)M_e(m,r)
		\end{array}\right].
	\end{equation*} 
    Since $F_r$ is stable, $N^{11}_c F_r N^{11}_e$ is stable. Moreover $M_c(m,r)M_e(m,r)=I_{\sigma(m,r)}$. This concludes the proof. 
\end{proof}

\addtolength{\textheight}{-12cm}   





\end{document}